\documentclass[preprint,12pt]{elsarticle}
\makeatletter
\def\ps@pprintTitle{%
 \let\@oddhead\@empty
 \let\@evenhead\@empty
 \def\@oddfoot{\centerline{\thepage}}%
 \let\@evenfoot\@oddfoot}
\makeatother

\usepackage{amssymb}
\usepackage{hyperref}

\usepackage[left=2cm, right=2cm, top=2cm]{geometry}
 
\usepackage{algorithmicx}
\usepackage{amsmath}
\usepackage{algorithm}
\usepackage{setspace}
\usepackage{lipsum}
\usepackage{mwe}
\usepackage{graphicx}
\usepackage{subcaption}
\usepackage[bottom]{footmisc}
\usepackage{graphicx}
\usepackage{algorithmicx}
\usepackage{algorithm}
\newcommand{\blank}[1]{\hspace*{#1}} 
\usepackage{setspace}
\usepackage{amsmath}

\usepackage{algpseudocode}
\usepackage{multirow}

\usepackage{algpseudocode}
\usepackage{booktabs}

\algnewcommand{\LineComment}[1]{\State \(\triangleright\) #1}

\usepackage[utf8]{inputenc}
\usepackage[english]{babel}
\usepackage{float}
\begin{document}

\begin{frontmatter}



\title{Localisation of Mammographic  masses  by  Greedy Backtracking of Activations in the Stacked Auto-Encoders.}


\author{*Shamna Pootheri (Research Fellow$^{a}$, Former Research Scholar$^{b}$)  and \\ V K Govindan  (Professor Emeritus $^{b}$) }

\address{$^{a}$Nanyang Technological University, Singapore-639798, $^{b}$ National Institute of Technology Calicut, India-673 601\\
 *Corresponding author: Email: shamnapootheri@gmail.com}
\begin{abstract}
Mammographic image analysis requires accurate localisation of salient mammographic masses. In mammographic computer-aided diagnosis,  mass or Region of Interest (ROI) is often marked by physicians and features are extracted from the marked ROI. In this paper, we present a novel mammographic mass localisation framework, based on the maximal class activations of the stacked auto-encoders.  We hypothesize that the image regions activating abnormal classes in mammographic images will be the breast masses which causes the anomaly. The experiment is conducted using randomly selected 200 mammographic images (100 normal and 100 abnormal) from IRMA mammographic dataset. Abnormal mass regions marked by an expert radiologist are used as the ground truth. The proposed  method outperforms existing  Deep Convolutional Neural Network (DCNN) based techniques in terms of salient region detection accuracy. The proposed  greedy backtracking method is more efficient and does not require a vast number of labelled training images as in DCNN based method. Such automatic localisation method will assist physicians to make accurate decisions on biopsy recommendations and treatment evaluations.\\


\end{abstract}


\end{frontmatter}
\section{Introduction }
Breast cancer is one of the noted reasons of cancer-related deaths among women. Mammographic image analysis is the widely accepted method for breast mass detection and diagnosis. The detection of mammographic masses is challenging due to the disparity of masses in shape, size, and low contrast from surrounding tissue. 

Deep learning techniques, especially transfer learning and fine-tuning methods, are often used for mammographic image analysis. Ribil et al.~\cite{ribli2018detecting} proposed an automatic mass localisation method using Faster R-CNN based techniques. Faster R-CNN has an additional layer called Region Proposal Network (RPN), on top of the last convolutional layer of the original network, which is trained to detect and localize salient image regions. The authors used VGG 16 pre-trained DCNN and fine-tuned the network with DDSM~\cite{heath2000digital} Mammographic images. Even though the network automatically detects lesions in the test image, the model required pixel-level annotated lesions to train the network and the performance of the system substantially  depends on the quality of the training images ~\cite{ribli2018detecting}. Rafael et al.~\cite{bressan2018doctor} proposed a deep transfer learning based mammographic patch retrieval framework. Features from the last convolution layers of the pre-trained DCNN (ResNet, Inception-V3, Nas-NetLarge) are used to estimate the similarity between images. The retrieval list is improved by a query refinement strategy using the Query-Point Movement approach (QPM). The performance of the retrieval system is compared with handcrafted and deep learning based features. The results of the initial iterations for both hands crafted and deep learned features are low. The precisions of deep learned features improve after a few iterations using query refinement. The performance of the method is evaluated using manually extracted ROI, and it is reported that the system requires multiple iterations with query refinement to obtain better results.

Recent  literature shows that a few works only are trying to interpret DCNN based features~\cite{zeiler2014visualizing,yosinski2015understanding,simonyan2013deep,springenberg2014striving} to locate salient regions  in  natural images. In~\cite{yosinski2015understanding}, class activation of the initial convolution layers of DCNN networks is used to visualize and locate the discriminating  image regions. Zeiler et al.~\cite{zeiler2014visualizing} occluded the input image by small patches, and the influence of each image regions on the class activation is observed to detect the significant image pixels. Simonyan et al.~\cite{simonyan2013deep} located salient image regions by back-propagating the gradients of selected neurons in each layer. During back-propagation, some paths have a positive impact, and some have a negative impact, and these paths combine to generate noisy gradients. Whereas, guided back-propagation~\cite{springenberg2014striving} method keeps the paths that lead to a  positive impact on the class score, and suppress the ones that have a negative impact, leading to better localisation. The localisation ability of the above methods is not yet analysed using medical images.

 As the DCNN used in the above techniques are trained with millions of natural images, we have to further analyse the performance of such systems in the medical domain. The main obstacle to use the DCNN  in the medical domain is the unavailability of large collection of images containing  annotated anomaly to train the network.   Moreover, deep-learned features are often described as black boxes since it is difficult to trace a prediction back to the features causing that prediction. Adapting such features in medical evaluation and diagnosis may have serious legal consequences. In this paper we try to  interpret the deep learned features to locate salient mammographic regions using unsupervised stacked auto-encoders.

The remaining part of this paper is presented as follows: Section 2 describes methods and algorithm of the proposed framework. Section 3 presents the experimental details and results of the proposed system. Section 4 provides comparative discussions of the results of the proposed method with state-of-the-art systems. Section 5 concludes the work.

\section{Material and Method }



\subsection{Mass Localisation by backtracking the maximally activated neurons}

 In mammographic computer-aided diagnosis, masses or Region of Interest (ROI) is often marked by physicians and features are extracted from the marked ROI~\cite{aminikhanghahi2017new,bressan2018doctor,li2018computer,purwadi2016assessment}.  In the proposed work we back-track the maximally activated neuron in each layer of a stacked auto-encoders to locate ROI in the mammographic images. An auto-encoder is an unsupervised artificial feed-forward neural network used to learn potential image features by reproducing the inputs from the hidden representations. We trained the  auto-encoders  consisting of one input layer, two hidden layers and one output layer to evaluate the performance of the proposed method.  Bias is ignored to find the direct influence of input pixels in activations of the neurons in each layer. Each layer in the auto-encoder is trained separately and combined to form the four layered stacked auto-encoder.

Let the size of the input images in the database be $M \times N$ pixels. The input layer accepts an input as the pixel intensities of  $\mathbf{x_i }= \{x_{ij },  j = \{1,2,…,M*N\}\}$. Where $\mathbf{x_i}$ represents $i^{th}$ input image and $x_{ij }$ represents intensity value of $j^{th}$ pixel in the $i^{th}$ image. The first hidden layer encodes the high-dimensional $(M*N)$ input image intensity to low dimensional (R) compressed representation  $\mathbf {h_1}$. The second hidden layer consisting of Q neurons find more interesting structures $\mathbf{ h_2}$   of the input data from $\mathbf{h_1}$. The output layer consisting of neuron proportions to the number of classes (C) present in the dataset. We used two neurons in the output layer representing normal and abnormal classes.

\begin{equation}
\label{eq3}
h_{1r}=f^e (\sum_{j=1}^J w_{rj } *  x_{ij})                                
\end{equation}

\begin{equation}
\label{eq4}
h_{2q}=f^e (\sum_{r=1}^R w_{qr } *  h_{1r})                                 
\end{equation}
\begin{equation}
\label{eq5}
z_{c}= (\sum_{q=1}^Q w_{cq } *  h_{2q})                               
\end{equation}

Where $ x_{ij} $  is the intensity of the   $j^{th}$ pixel in the $ i^{th}$ image,
$J$ is the number of neurons in the input layer,
 $R$ is the number of neurons the hidden layer one,
 $Q$ is the number of neuron in the hidden layer two,
$C$ is the number of neurons in the output layer,
 and $J>R>Q>C$.

The ${h_{1r}}$ represents the activation of $r^{th}$ neuron in the hidden layer one (Equation \ref{eq3}), $w_{rj}$ represents the weight between the $r^{th}$ neuron in the hidden layer one and $j^{th}$ input in the input layer (  $w_{rj}$ signifies the importance of $j^{th}$ image pixel in activating the $r^{th}$ neuron in the hidden layer one). Similarly $h_{2q}$, $ w_{qr}$  ( in Equation \ref{eq4}) are the class activation and weights of $q^{th}$ neuron in hidden layer two; and $z_c$, $w_{cq }$  ( in Equation \ref{eq5}) are the class activation  and weights of output layer. We used sigmoid activation function ($ f^e$ )  given in Equation \ref{eq6} to encode and decode the data. The results from the output layer ($ z_c$) is passed to a Soft-max function (Equation \ref{eq7}). Soft-max function produces the probability ($P_c$  )  of the input image in each class. Based on the probability the input image is classified to normal or abnormal class. 

\begin{equation}
\label{eq6}
Sigmoid(a ) =\frac{\exp(a) }{ \exp(a)+1 }                              
\end{equation}

\begin{equation}
\label{eq7}
P_c=Softmax(z_c ) =\frac{\exp(z_c) }{\sum_{c=1}^C \exp(z_c) }                              
\end{equation}

\begin{algorithm}\label{greedy}
\caption{:Greedy Backtracking.}\label{euclid}
 \textbf {\\Input :} $\mathbf{{<x_i}}$\textbf {, AE, S}$\mathbf{ >}$

 $\mathbf{x_i}=\{x_{ij } \mid j = \{1,2,…,M*N\}\} $  \\     \Comment{\parbox[t]{1\linewidth}{{\footnotesize $\mathbf{ x_i}$ be the $i^{th}$ image in the database, $ x_{ij}$ be the intensity of the $j^{th}$  pixel in  image   $\mathbf{ x_i} $  }}}\\ 
 $ \textbf{AE}(InputLayer, HiddenLayer1,HiddenLayer2,OutputLayer)$\\
 \Comment{\parbox[t]{.95\linewidth}{{\footnotesize Stacked auto-encoders with one input layer, two hidden layers and  one output layer }}}
\\Parameter's of \textbf{AE}:
         \\  J \Comment{\parbox[t]{.9\linewidth}{{\footnotesize  Number of neurons in the $InputLayer$, J=M*N}}}
         \\ R \Comment{\parbox[t]{.9\linewidth}{{\footnotesize  Number of neurons in the $HiddenLayer1$}}} 
         \\Q\Comment{\parbox[t]{.9\linewidth}{{\footnotesize   Number of neurons in the $HiddenLayer2$}}} 
         \\  C\Comment{\parbox[t]{.9\linewidth}{{\footnotesize   Number of neurons in the $OutputLayer$/Number of Classes}}}
\\ $w_{rj}$ \Comment{\parbox[t]{.9\linewidth}{{\footnotesize   Weight between the $j^{th}$ neuron in the $InputLayer$ and $r^{th}$ neuron in the $HiddenLayer1$}}}
\\ $h_{1r}$ \Comment{\parbox[t]{.9\linewidth}{{\footnotesize   Output from the $r^{th}$ neuron in the $HiddenLayer1$}}}
\\ $w_{qr}$ \Comment{\parbox[t]{.9\linewidth}{{\footnotesize   Weight between the $r^{th}$ neuron in the $HiddenLayer1$ and $q^{th}$ neuron in the $HiddenLayer2$}}}
\\ $h_{2q}$\Comment{\parbox[t]{.9\linewidth}{{\footnotesize   Output from the $q^{th}$ neuron in the $HiddenLayer2$}}}
\\ $w_{cq }$  \Comment{\parbox[t]{.9\linewidth}{{\footnotesize  Weight between the $q^{th}$ neuron in the $HiddenLayer2$ and $c^{th}$ neuron in the $OutputLayer$}}}\\
\\\textbf{S :} Count of Salient pixels \\ \Comment{\parbox[t]{.63\linewidth}{{\footnotesize  Number of pixels considered for seed point localisation}}}    \\             
 \textbf  {Output : } $\mathbf{<}$\textbf{  SP($c_x,c_y$)}$\mathbf{>} $\\ 
   {\textbf{  SP($c_x,c_y$):} Seed Point Coordinates}\\ \Comment{\parbox[t]{.75\linewidth}{{\footnotesize Image cordinate representing the seedpoint to locate anomaly.}}}

\noindent\rule[0.5ex]{\linewidth}{1pt} \\
\begin{algorithmic}[1]
\setstretch{1}
\State  	Input  $ \mathbf{x_i }$ in to the \textbf{AE} and find probability of the image in each class :$\{ P(c )\mid c\in \{1, 2,...,C \}\} $ 
\State{Find the   class of the images based on the maximum probability of Softmax function in the output layer. }	
\Statex  \blank{1.5cm}	$c =  \underset{c}{\mathrm{argmax}}\{ P(c )\mid  c\in \{1, 2,...,C \}\}$ \Comment{\parbox[t]{.35\linewidth}{{\footnotesize Let the class of image be $c$  }}}	
\State {Backtrack  the $c^{th}$ neuron in  $OutputLayer$ and find the neuron in   $ HiddenLayer2$ which causes maximum  activation in the predicted class $c$.}	
\Statex 	\blank{2.5cm} $q =  \underset{q}{{argmax}}$$\{ (w_{cq}* h_{2q}) \mid  q\in \{1, 2,...,Q\}\}$ 
\Statex  {  \Comment{\parbox[t]{.8\linewidth}{{\footnotesize  Let  $q$  be the back-tracked neuron in $ HiddenLayer2$  }}}}

\State {Backtrack the $q^{th}$ neuron in  $HiddenLayer2$  and find the neuron in $HiddenLayer1$  which causes max activation in the $q^{th}$ neuron.}	
\Statex 	\blank{2.5cm} $r =  \underset{r}{\mathrm{argmax}}\{ (w_{qr}* h_{1r}) \mid  r\in \{1, 2,...,R\}\}$  
\Statex{  \Comment{\parbox[t]{.8\linewidth}{{\footnotesize  Let  $r$  be the back-tracked neuron in $ HiddenLayer1$  }}}}
\State Sort the activations of the  $r^{th}$ neuron in the $HiddenLayer1$ in descending order.
\Statex 		 $SortedActivations = Sort Descending\{(w_{rj} *x_{ij }) \mid  j=\{1,2,.., MXN\} \}  $ 
\State Take top 'S' image pixels ( $x_{ij}$)  from $SortedActivations$ which causes highest class activations in the  $r^{th}$ neuron in the HiddenLayer1.
\State Collect the coordinates of the selected top ‘S’ pixels in the input image  $ \mathbf{x_i }$
\State Cluster the coordinate in the denser region and eliminate the outliers.
\State The cluster centre cordinates  represents the  seedpoint  ($\textbf{  SP($c_x,c_y$)}$) to locate anomaly.

\end{algorithmic}
\end{algorithm}

The proposed mass localisation procedure by backtracking is listed in Algorithm 1. An input image is fed in to the stacked auto-encoders with one input layer, two hidden layers and one output layer. The image class is predicted based on the probability of the soft-max layer. The output layer neuron corresponding to the predicted class is back-tracked and the neuron in hidden layer 2, which causes the highest activation in the predicted class is identified.  Similarly, the layer 1 neuron which causes maximal activation is pinpointed, by backtracking the identified neuron in the layer 2. After locating the neuron in layer 1, the coordinates of the top ‘S’ image pixels (Salient image pixels) causing maximum activation in the located layer1 neuron are selected. The selected co-ordinates in the denser regions are clustered and outliers are removed. If more than one cluster is present the cluster centre with highest pixel intensity is selected as the seed point coordinates of the given image (since the abnormal masses have higher intensity value compared to normal masses). Figure~\ref{f2} represents a pictorial overview of the stacked auto-encoder and the backtracking mechanism to find seed point location in the input image. After locating the seed point, the ROI is extracted and segmented automatically using the region growing algorithm~\cite{adams1994seeded}.

\begin{figure}\label{f2}
\includegraphics[width=1\textwidth]{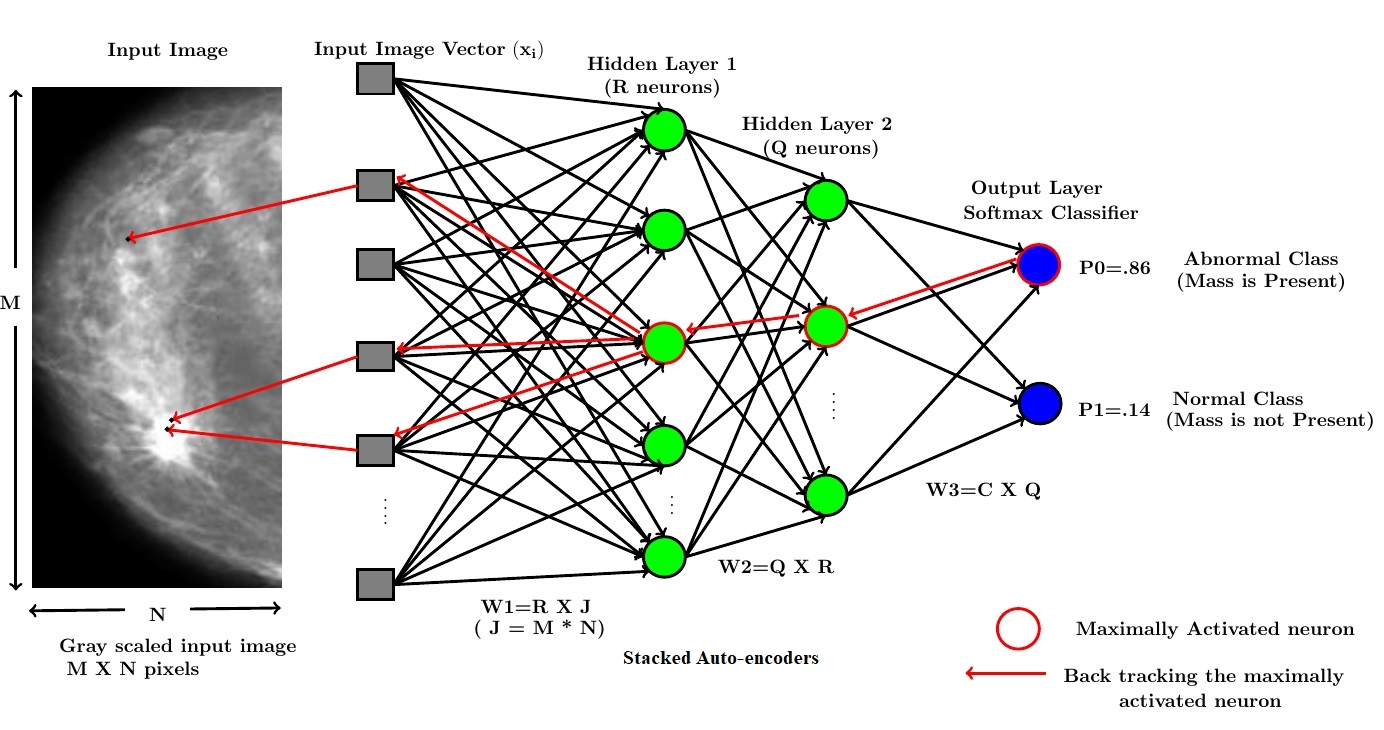}
\caption{(colour)The pictorial overview of the  backtraking the trained autoencoder to locate salient image pixels in the input image .} \label{f2}
\end{figure}
\pagebreak
\section{Experimental Details and Results} The performance of the proposed framework is assessed using randomly selected 200 images from IRMA mammographic dataset (100 normal mammograms and 100 abnormal images with mammographic masses are selected). Seventy percentage of the images from each class are used to train the auto-encoder.  All images are resized to $256\times128$ pixels size, and the pectoral muscle area is removed\cite{camilus2010computer}. The mammographic masses are marked by an expert radiologist, which is used as the ground truth for the experiment. The proposed method is implemented using Matlab 2018b software in a system with 8 GB RAM and Intel i7 6500U CPU. The DCNN based ROI  localisation is performed using~\cite{grun2016taxonomy}. We trained a stacked auto-encoders with one input layer, two hidden layer and one output layer. The input layer consists of neuron corresponding to the image size (J=$256\times128$) and the output layer consists of neuron corresponding to the number of classes (C=2, representing normal and abnormal classes). The number of neuron in the hidden layer 1 (R=100), the number of neuron in the hidden layer 2 (Q=10), and the salient image pixels (S=20) is determined empirically.

\begin{figure}
\begin{center}
\includegraphics[width=.5\textwidth]{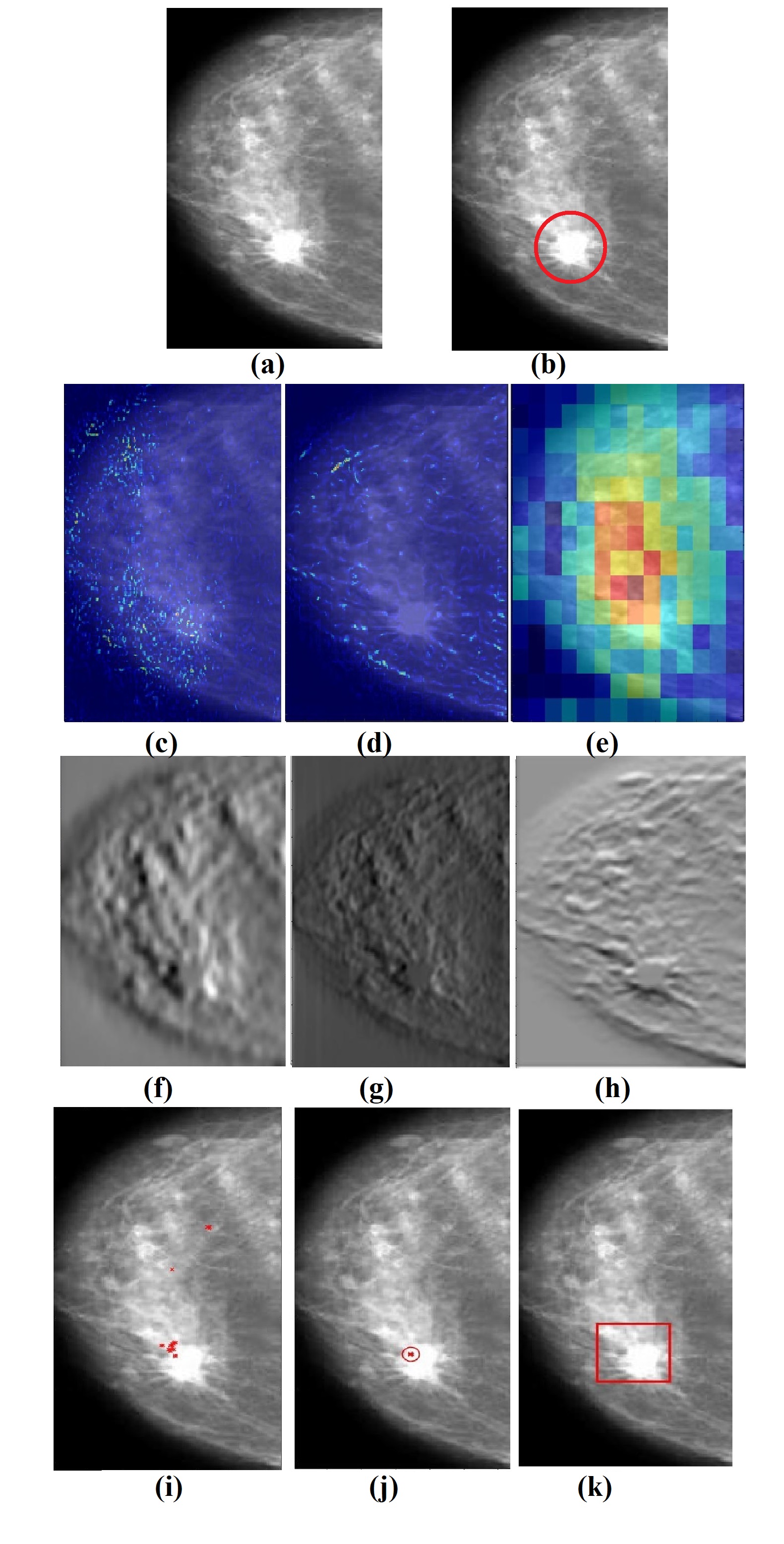}
\caption{(colour)\textbf{(a)} A sample mammographic image. \textbf{(b)} ROI marked by the  radiologist. \textbf{(c)} ROI located by  Back propogation~\cite{simonyan2013deep} \textbf{(d)} ROI located by Guided back propogration~\cite{springenberg2014striving} \textbf{(e)}  ROI located by  Occlusion method~\cite{zeiler2014visualizing}. [Red colour pixels indicates salient regions in \textbf{(c-e )}],\textbf{ (f) }Class activations of Alexnet~\cite{krizhevsky2012imagenet}. \textbf{(g)} Class activations of Googlnet~\cite{szegedy2015going}. \textbf{(h)} Class activations of Resnet  \cite{he2016deep}.[  High pixel intensity(white colour or light coloured pixels)  indicates salient regions in class activatios of \textbf{(f-h)}], \textbf{(i)} the proposed method (Salient 20 pixels are represented in red colour). \textbf{(j)} Selected seed point from the salient pixels \textbf{(k)} ROI extracted using region growing algorithm~\cite{adams1994seeded} represented  in  the red bounding box }  \label{fig2}
\end{center}
\end{figure}

We evaluated the mass localisation ability of DCNN in the mammographic images using Occlusion~\cite{zeiler2014visualizing}, Back-Propogation~\cite{simonyan2013deep} and Guided Back-Propogation~\cite{springenberg2014striving} method using pre-trained GoogleNet~\cite{szegedy2015going}. We also analysed the localisation ability of fine-tuned DCNN ( Alexnet~\cite{krizhevsky2012imagenet}, GoogleNet~\cite{szegedy2015going} and Resnet~\cite{he2016deep}), by examine the maximum class  activation in the initial convolutional layers~\cite{yosinski2015understanding}. The ROI marked by the radiologist (100 mammographic images in the abnormal class with one suspicious area) is used as the ground truth in the mass localisation experiment. Table \ref{tab1} shows the count of correctly located and wrongly located masses in the mammographic images by the proposed method and DCNN based method~\cite{zeiler2014visualizing,yosinski2015understanding,simonyan2013deep,springenberg2014striving}. Figure \ref{fig2} depicts a mammographic image with located masses using the proposed method and DCNN based methods. The results in Table \ref{tab1} and Figure \ref{fig2} clearly shows that the proposed method has good localisation ability to detects masses in mammographic images compared to the DCNN based methods.


\begin{table}[h!]
\renewcommand{\arraystretch}{1.13}
 \caption{Mass localisation in 100 abnormal Mammograms}
 \centering
 \scalebox{0.7}{

\begin{tabular}{lcc}
  \hline 
\textbf{Method}  & \textbf{\#Correctly Located} & \textbf{\#Wrongly Located} \\
  \hline 
{\textbf{Proposed Method III}} & {\textbf{87}}    &{\textbf{13}}              \\
AlexNet (2012) \cite{krizhevsky2012imagenet}                 & 65                         & 35                       \\
Occlusion (2014) \cite{zeiler2014visualizing}         & 58                         & 42                       \\
GoogleNet(2015) \cite{szegedy2015going}            & 34                         & 66                       \\
ResNet (2016) \cite{he2016deep}              & 27                         & 73                       \\
Guided back-Propagation (2014) \cite{springenberg2014striving} & 19                         & 81                       \\
Back-Propagation  (2014) \cite{simonyan2013deep}     & 17                         & 83              \\     
 \hline 
 \end{tabular}}
\label{tab1}
\end{table}
\pagebreak

We also compared the performance of  the proposed method with an automatic mass localisation method using Faster R-CNN proposed by Ribli et al.~\cite{ribli2018detecting}.  We used the   CnnCAD~\cite{CADCnn}  plug-in provided  by the author for evaluating the performance. The CnnCAD plug-in is implemented in Horos in a  mac computer with 4 GB RAM and High Sierra (10.13) operating system. Table~ \ref{tab2} provides the localisation performance of the proposed method with Faster R-CNN proposed by the Ribli et al.~\cite{ribli2018detecting}. The  experimental results in Table~\ref{tab2} indicate  that the performance of supervised Faster R-CNN~\cite{ribli2018detecting} detecting system  is not promising  since its performance is substantially depends on the quality of the training images~\cite{ribli2018detecting}. Figure~ \ref{fig3} provides the samples of ROI detected by the proposed method and Faster R-CNN ~\cite{ribli2018detecting}. In Faster R-CNN~\cite{CADCnn}  method ROI markers will appear automatically on the image, if the model detects any abnormal image region.

\begin{table}[h!]
\renewcommand{\arraystretch}{1.13}
 \caption{Performance comparison of the proposed method with Faster-RCNN method}
 \centering
 \scalebox{0.7}{

\begin{tabular}{lcc}
  \hline 
\textbf{Method}  & \textbf{\#Correctly Located} & \textbf{\#Wrongly Located} \\
  \hline 
{\textbf{Proposed Method III}} & {\textbf{87}}    &{\textbf{13}}              \\

Faster R-CNN  (2014) ~\cite{ribli2018detecting}     & 11                         & 89              \\     
 \hline 
 \end{tabular}}
\label{tab2}
\end{table}
\begin{figure}\label{fig3}
\includegraphics[width=1\textwidth]{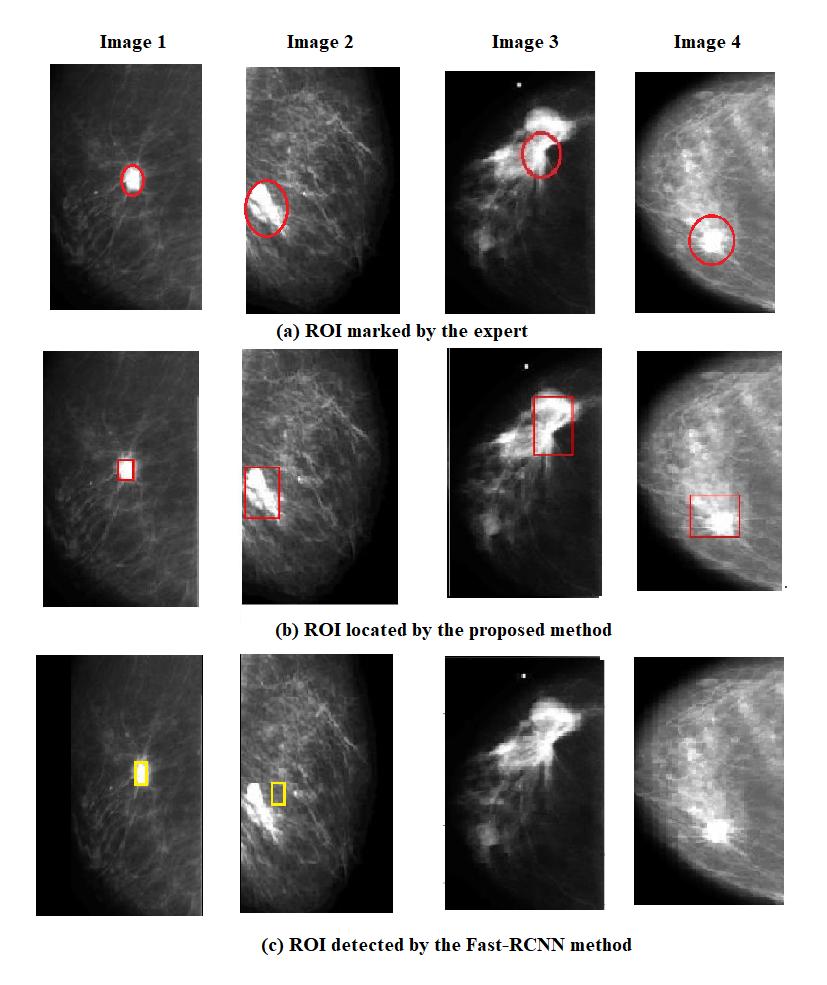}
\caption{(colour) ROI detected by the proposed method and Faster R-CNN~\cite{ribli2018detecting}  in four sample images. First row (a) Represents the ROI marked by the expert. Second row (b) Indicates the ROI located by the proposed method. Third Row (c) Represents the ROI detected by the Faster R-CNN method~\cite{ribli2018detecting}.} \label{fig3}
\end{figure}
\pagebreak
\section{Discussion} 

The overall results show that the proposed greedy backtracking method is less complex compared to the DCNN based method, and more efficient to locate salient mammographic masses as shown in  Table \ref{tab1} and \ref{tab2}. The performance of the DCNN based methods heavily depends on the quality and quantity of the labelled images used to train the network~\cite{cho2015much,tajbakhsh2016convolutional}. Most of the medical images are normally noisy due to the imaging conditions and the labelled images are not available at affordable cost in the medical field. The auto-encoder requires a limited number of unlabelled training images to de-noise and represent the input image efficiently compared to DCNN~\cite{chen2017deep}. The convolutional layers in DCNN~\cite{yosinski2015understanding} effectively capture discriminating colour and texture features of natural images to detect salient image regions.  Most of the medical images are grey coloured and do not have sharp edged features as in natural images. The proposed greedy backtacking method using auto-encoders can detect ROI in grey-coloured mammograms very efficiently, compared to state-of-the-art DCNN based methods~\cite{ribli2018detecting, simonyan2013deep,springenberg2014striving,yosinski2015understanding,zeiler2014visualizing}. 

The experimental results in Figure~\ref{fig3} show that, the Faster R-CNN~\cite{CADCnn}  method is not able to detect the ROI’s in the mammographic images with significant mammographic masses (Figure~\ref{fig3}: Image 2 ROI  is wrongly detected,  Figure~\ref{fig3}: Image 3 and 4 not detected any ROI). This may be mainly due to the image quality requirement for Faster R-CNN~\cite{ribli2018detecting}. However, the proposed method is able to detect the ROI’s with the same quality images. Moreover, Faster R-CNN~\cite{ribli2018detecting} required pixel level annotated images to train the network, whereas the proposed back-tracking method using unsupervised auto-encoder doesn’t require any labelled data.


In this paper,  we hypothesized that the image regions activating abnormal classes in mammographic images will be the breast masses which causes the anomaly and back-tracked the maximal activations of stacked auto-encoders to locate ROI. Such interpretations can help to promote the acceptance of deep learned features in clinical evaluation and patient treatment process. As it is difficult to attain large collection of medical images with unusual masses, our aim was to analyse the localisation ability of the deep network with a limited number of images.  The unsupervised auto-encoders do not require a huge number of  labelled training images, so the proposed method  is apt for medical image application. Additionally, the proposed method  is computationally less complex in locating salient image regions compared to the DCNN based methods~\cite{ribli2018detecting,zeiler2014visualizing,yosinski2015understanding,simonyan2013deep,springenberg2014striving}.  We have to further analyse the performance of the system using  more complex dataset, containing mammographic images with multiple  suspicious masses.
  
\section{Conclusion} 

Mass localisation in the mammographic images is the critical step in computer-aided mammographic analysis methods. In this paper, we have proposed an automatic mammographic mass localisation and retrieval framework using stacked auto-encoders. The proposed method back-tracks the maximal class activations in the stacked auto-encoders. The DCNN based methods have poor localisation ability due to the absence of coloured and sharp edged features in medical images. The experimental results demonstrated that the proposed method is superior in localising mammographic images with a limited number of unlabelled medical images compared to DCNN based methods.

 \bibliographystyle{elsarticle-num} 
 \bibliography{References1}
\end{document}